\documentclass[11pt,letterpaper]{article}
\usepackage{graphicx}
\graphicspath{ {images/} }
\usepackage{emnlp2017}
\usepackage{times}
\usepackage{latexsym}

\def\realnumbers{\rm I\!R}

\title{Do Convolutional Networks need to be Deep for Text Classification ?}
\author{Hoa T. LE \textsuperscript{1}, Christophe Cerisara \textsuperscript{1}, Alexandre Denis \textsuperscript{2} \\
  {\tt \textsuperscript{1} LORIA, UMR 7503, Nancy, France; \textsuperscript{2} SESAMm, France}}

\date{}

\begin{document}

\maketitle

\begin{abstract}

We study in this work the importance of depth in convolutional models for text classification, either when character or word inputs are
considered. We show on 5 standard text classification and sentiment analysis tasks that 
deep models indeed give better performances than shallow networks when the text input is represented as a sequence of characters.
However, a simple shallow-and-wide network outperforms deep models such as DenseNet with word inputs.
Our shallow word model further establishes new state-of-the-art performances on two datasets: Yelp Binary (95.9\%) and Yelp Full (64.9\%).
\end{abstract}

\section{Introduction}

Following the success of deep learning approaches in the ImageNet competition, there has been a surge of interest in the application of deep models on 
many tasks, such as image classification~\cite{Krizhevsky-et-al:2012}, speech recognition~\cite{Hinton-et-al:2012}, ...
Each new coming model proposes better and better architectures of the network to facilitate training through longer and longer chains of layers,
such as Alexnet~\cite{Krizhevsky-et-al:2012}, VGGNet~\cite{Simonyan-and-Zisserman:2014}, GoogleLeNet~\cite{Szegedy-et-al:2015}, ResNet~\cite{He-et-al:2015}
and more recently Densenet~\cite{Huang-et-al:2016}.

Several works have explained this success in computer vision, in particular~\cite{Zeiler-and-Fergus:2014} and~\cite{Donahue-et-al:2014}: deep model is able to learn a hiearchical feature representation from pixels to line, contour, shape and object.
These studies have not only helped to demistify the black-box of deep learning, but have also led the path to other approaches like transfer learning~\cite{Yosinski-et-al:2014},
where the first layers are believed to bring more general information and the last layers to convey specific information on the target task. 

This paradigm has also been applied in text classification and sentiment analysis, with deeper and deeper networks being proposed in the literature:
\cite{Kim:2014} (shallow-and-wide CNN layer), \cite{Zhang-et-al:2015} (6 CNN layers), \cite{Conneau-et-al:2016} (29 CNN layers).
Besides the development of deep networks, there is a debate about which atom-level (word or character) would be the most effective for Natural Language Processing (NLP) tasks.
Word embeddings, which are continuous representations of words, initially proposed by~\cite{Bengio-et-al:2003} and widely adopted after word2vec~\cite{Mikolov-et-al:2013}
have been chosen as the main standard representations for most NLP tasks.
Based on this representation, a common belief is that, similarly to vision, the model will learn hiearchical features from the text: words combine to form n-grams, phrases, sentences...

Several recent works have extended this model to characters instead of words. Hence, \cite{Zhang-et-al:2015} propose for the first time an alternative character-based model,
while \cite{Conneau-et-al:2016} take a further step by introducing a very deep char-level network.
Nevertheless, it is still not clear which atom-level is the best and whether very deep networks at the word-level are really better for text classification.

This work is motivated by these questions and we hope to bring elements of a response
by providing a full comparison of a shallow-and-wide CNN~\cite{Kim:2014} both at the character and word levels on the 5 datasets described in~\cite{Zhang-et-al:2015}.
Moreover, we propose an adaptation of a DenseNet~\cite{Huang-et-al:2016} for text classification and sentiment analysis at the word-level, which we compare with the state-of-the-art.

This paper is structured as follows:
Section~\ref{sec:sota} summarizes the related work while
Section~\ref{sec:model} describes the shallow CNN and introduces our adaptation of DenseNet for text.
We then evaluate our approach on the 5 datasets of~\cite{Zhang-et-al:2015} and show experimental results in Section~\ref{sec:xp}.
Experiments show that the shallow-and-wide CNN on word-level can beat a very deep CNN on char-level.
The paper concludes with some open discussions for future research about deep structures on text.

\section{Related work}
\label{sec:sota}

Text classification is an important task in Natural Language Processing. Traditionally, linear classifiers are often used for text classification~\cite{Joachims:1998, McCallum:1998, Fan-et-al:2008}. In particular, \cite{Joulin-et-al:2016} show that linear models could scale to a very large dataset rapidly with a proper rank constraint and a fast loss approximation.
However, a recent trend in the domain is to exploit deep learning methods, such as convolutional neural networks:~\cite{Kim:2014, Zhang-et-al:2015, Conneau-et-al:2016} and recurrent networks:~\cite{Yogatama-et-al:2017, Xiao-and-Cho:2016}.
Sentiment Analysis is also an active topic of research in NLP for a long time, with real-world applications in
market research~\cite{Qureshi-et-al:2013}, finance~\cite{Bollen-et-al:2011}, social science~\cite{Dodds-et-al:2011}, politics~\cite{Kaya-et-al:2013}.
The SemEval challenge has been setup in 2013 to boost this field and is still bringing together many competitors
who have been using an increasing proportion of deep learning models over the years~\cite{Nakov-et-al:2013,Rosenthal-et-al:2014,Nakov-et-al:2016}.
2017 is the fifth edition of the competition, with at least 20 teams (over 48 teams) using deep learning and neural network methods.
The top 5 winning teams all use deep learning or deep learning ensembles.
Other teams use classifiers such as Naive Bayes classifier, Random Forest, Logistic Regression, Maximum Entropy and Conditional Random Fields~\cite{Rosenthal-et-al:2017}.

Convolutional neural networks with end-to-end training have been used in NLP for the first time in~\cite{Collobert-and-Weston:2008,Collobert-et-al:2011}.
The authors introduce a new \textit{global} max-pooling operation, which is shown to be effective for text, as an alternative to the conventional \textit{local} max-pooling
of the original LeNet architecture~\cite{Lecun-et-al:1998}.
Moreover, they proposed to transfer task-specific information by co-training multiple deep models on many tasks.
Inspired by this seminal work, \cite{Kim:2014} proposed a simpler architecture with slight modifications of~\cite{Collobert-and-Weston:2008} 
consisting of fine-tuned or fixed pretraining word2vec embeddings~\cite{Mikolov-et-al:2013} and its combination as multi-channel.
The author showed that this simple model can already achieve state-of-the-art performances on many small datasets.
\cite{Kalchbrenner-et-al:2014} proposed a dynamic \textit{k}-max pooling to handle variable-length input sentences.
This dynamic \textit{k}-max pooling is a generalisation of the max pooling operator where \textit{k} can be dynamically set as a part of the network. 

All of these works are based on word input tokens, following~\citep{Bengio-et-al:2003}, which introduced for the first time a solution to fight the curse of dimensionality
thanks to distributed representations, also known as \textit{word embeddings}.
A limit of this approach is that typical sentences and paragraphs contain a small number of words, which prevents the previous convolutional models to be very deep: most of them indeed only have two layers.
Other works~\cite{Santos-and-Gatti:2014} 
further noted that word-based input representations may not be very well adapted to social media inputs like Twitter, 
where tokens usage may be extremely creative: slang, elongated words, contiguous sequences of exclamation marks, abbreviations, hashtags,...
Therefore, they introduced a convolutional operator on characters to automatically learn the notions of words and sentences.
This enables neural networks to be trained end-to-end on texts without any pre-processing, not even tokenization.
Later, \cite{Zhang-et-al:2015} enhanced this approach and proposed a deep CNN for text:
the number of characters in a sentence or paragraph being much longer, they can train for the first time up to 6 convolutional layers.
However, the structure of this model is designed by hand by experts and it is thus difficult to extend or generalize the model with arbitrarily different kernels and pool sizes.
Hence, \cite{Conneau-et-al:2016}, inspired by \cite{He-et-al:2016}, presented a much simpler but very deep model with 29 convolutional layers. 

Besides convolutional networks, \cite{Kim-et-al:2016} introduced a character aware neural language model by combining a CNN on character embeddings with an highway LSTM on subsequent layers.
\cite{Radford-et-al:2017} also explored a multiplicative LSTM (mLSTM) on character embeddings and found that a basic logistic regression learned on this representation can achieve
state-of-the-art result on the Sentiment Tree Bank dataset \cite{Socher-et-al:2013} with only a few hundred labeled examples.

Capitalizing on the effectiveness of character embeddings, \cite{Dhingra-et-al:2016} proposed a hybrid word-character models to leverage the avantages of both worlds.
However, their initial experiments show that this simple hybridation does not bring very good results:
the learned representations of frequent and rare tokens of words and characters is different and co-training them may be harmful.
To alleviate this issue, \cite{Miyamoto-and-Cho:2016} proposed a scalar gate to control the ratio of both representations,
but empiricial studies showed that this fixed gate may lead to suboptimal results.
\cite{Yang-et-al:2016} then introduced a fine-grained gating mechanism to combine both representations.
They showed improved performance on reading comprehension datasets, including Children's Book Test and SQuAD.

\section{Model}
\label{sec:model}

We describe next two models architectures, respectively shallow and deep, that we will compare in Section~\ref{sec:xp} on several text classifications tasks.
Both models share common components that are described next.

\subsection{Common components}

\subsubsection*{Lookup-Table Layer}

Every token (either word or character in this work) $\textit{i} \in Vocab$ is encoded as a \textit{d}-dimensional vector using a lookup table $Lookup_W (.)$:
\begin{equation}
Lookup_W (i) = \textbf{W}_i,
\end{equation}

where $\textbf{W} \in \realnumbers^{\textit{d} \times \vert \textit{Vocab} \vert}$ is the embedding matrix, $\textbf{W}_i \in \realnumbers^{d}$ is the $i^{th}$ column of \textbf{W} and ${d}$ is the number of embedding space dimensions.
The first layer of our model thus transforms indices of an input sentence ${{s_1}, {s_2}, \cdots, {s_n}}$ of \textit{n} tokens in \textit{Vocab} into a series of vectors   
${{\textbf{W}_{s1}}, {\textbf{W}_{s2}}, \cdots, {\textbf{W}_{sn}}}$.

\noindent{\bf Classification Layer}

The embedding vectors that encode a complete input sentence are processed by one of our main models, which outputs
a feature vector $\textbf{x}$ that represents the whole sentence.
This vector is then passed to a classification layer that applies a \textit{softmax} activation function~\cite{Bridle:1990} to compute the predictive probabilities for all $K$ target labels:
\begin{equation}
p(y=k \vert X) = \frac{exp(\textbf{w}_k^T\textbf{x}+b_k)}{\sum_{k^{'}=1}^K exp(\textbf{w}_{k^{'}}^T\textbf{x}+b_{k^{'}})}
\end{equation}
where the weight and bias parameters $\textbf{w}_k$ and ${b}_k$ are trained simultaneously with the main model's parameters. The loss function is then minimized by cross-entropy error.

\subsection{Shallow-and-wide CNN}

Our first shallow-and-wide CNN model is adapted from~\citep{Kim:2014}.
\begin{figure}[h!]
\centering
\includegraphics[scale=0.4]{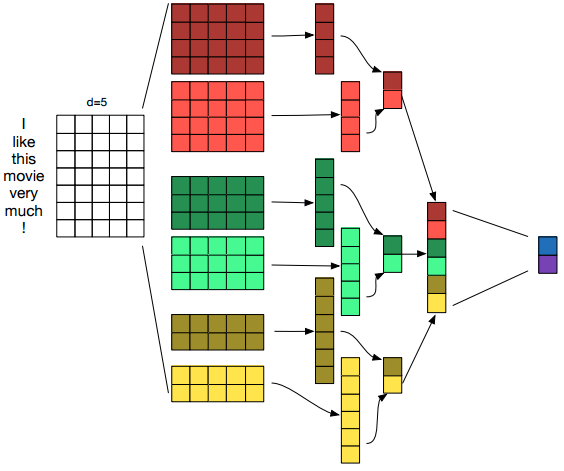}
\caption{Shallow-and-wide CNN, from~\protect\cite{Zhang-and-Wallace:2015}: 3 convolutional layers with respective kernel window sizes 3,4,5 are used. A global max-pooling is then applied to the whole sequence on each filter. Finally, the outputs of each kernel are concatenated to a unique vector and fed to a fully connected layer.}
\end{figure}

Let $\textbf{x}_i \in \realnumbers^d$ be an input token (\textit{word} or \textit{character}).
An input \textit{h}-grams $x_{i:i+h-1}$ is transformed through a convolution filter $\textbf{w}_c \in \realnumbers^{hd}$:
\begin{equation}
c_i = f(\textbf{w}_c \cdot \textbf{x}_{i:i+h-1} + b_c)
\end{equation}
with $b_c \in \realnumbers$ a bias term and \textit{f} the non-linear ReLU function.
This produces a \textit{feature map} $\textbf{c} \in R^{n-h+1}$, where $n$ is the number of tokens in the sentence. Then we apply a global max-over-time pooling over the feature map:
\begin{equation}
\hat{c} = \max\{\textbf{c}\} \in \realnumbers
\end{equation}

This process for one feature is repeated to obtain $m$ filters with different window sizes $h$. The resulting filters are concatenated to form a \textit{shallow-and-wide} network:
\begin{equation}
\textbf{g} = [ \hat{c}_1, \hat{c}_2, \cdots, \hat{c}_m]
\end{equation}

Finally, a fully connected layer is applied:
\begin{equation}
\hat{y} = f(\textbf{w}_y \cdot \textbf{g} + b_y)
\end{equation}

\subsubsection*{Implementation Details}

The kernel window sizes $h$ for character tokens are $N_f=(15,20,25)$ with $m=700$ filters.
For word-level, $N_f=(3,4,5)$ with $m=100$ filters.

\subsection{DenseNet}

\begin{figure}[h!]
\centering
\includegraphics[scale=0.6]{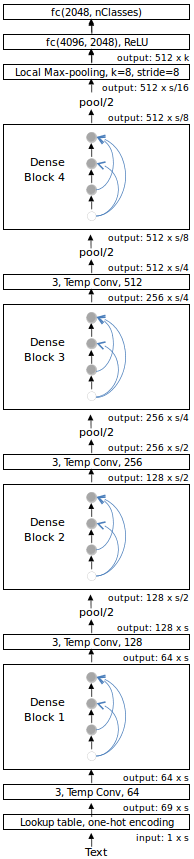}
\caption{Character-level DenseNet model for Text classification. \textbf{3, Temp Conv, 128} means temporal convolutional operation with kernel window size = 3 and filter size = 64; \textbf{pool/2} means local max-pooling with kernel size = stride size = 2, it will reduce the size of the sequence by a half.}
\label{DenseNet}
\end{figure}

\subsubsection*{Skip-connections}

In order to increase the depth of deep models, \citep{He-et-al:2015} introduced a skip-connection that modifies the non-linear transformation
$\textbf{x}_l = \mathcal{F}_l (\textbf{x}_{l-1})$ 
between the output activations $\textbf{x}_{l-1}$ at layer $l-1$ and at layer $l$
with an identity function:
\begin{equation}
\textbf{x}_l = \mathcal{F}_l (\textbf{x}_{l-1}) + \textbf{x}_{l-1}
\end{equation}

This allows the gradient to backpropagate deeper in the network and limits the impact of various issues such as vanishing gradients.

\subsubsection*{Dense Connectivity}

\cite{Huang-et-al:2016} suggested that the additive combination of this skip connection with $\mathcal{F}_l (\textbf{x}_{l-1})$ may negatively affect the information flow in the model.
They proposed an alternative concatenation operator, which allows to create direct connections from any layer to all subsequent layers, called \textit{DenseNet}.
Hence, the ${l^{th}}$ layer has access to the feature maps of all preceding layers, $\textbf{x}_0,\cdots,\textbf{x}_{l-1}$, as input:
\begin{equation}
\textbf{x}_l = \mathcal{F}_l ([\textbf{x}_0,\textbf{x}_1,\cdots,\textbf{x}_{l-1}])
\end{equation}

\begin{figure}[h!]
\centering
\includegraphics[scale=0.5]{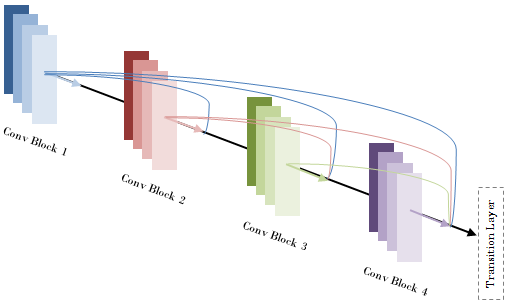}
\caption{Dense Block. Multiple convolutional filters output 2D matrices, which are all concatenated together before going into another dense block.}
\label{DenseBlock}
\end{figure}

This can be viewed as an extreme case of a ResNet. The distance between both ends of the network is shrinked and the gradient may backpropagate more easily from the output back to the input, as illustrated in Figure~\ref{DenseBlock}.

\subsubsection*{Convolutional Block and Transitional Layer}

Following~\cite{He-et-al:2016}, we define $\mathcal{F}_l (.)$ as a function of three consecutive operations: batch normalization (BN), rectified linear unit (ReLU) and a 1x3 convolution.

To adapt the variability of the changing dimension of the concatenation operation, we define a transition layer which composes a 1x3 convolution and a 1x2 local max-pooling between two dense blocks. Given a vector $c^{l-1}$ outputed by a convolutional layer $l-1$, the local max-pooling layer \textit{l} outputs a vector $c^{l}$:

\begin{equation}
\big[c^{l}\big]_j = \max_{i} \big[c^{l-1}\big]_{k \times (j-1) \leq i < k \times j}
\end{equation}

where $1\leq i\leq n$ and \textit{k} is the kernel pooling size. The word-level DenseNet model is the same as the character-level model shown in Figure~\ref{DenseNet}, except for the last two layers,
where the local max-pooling and two fully connected layers are replaced by a single global average pooling layer.
We empirically observed that better results are thus obtained with word tokens.

\subsubsection*{Implementation Details}

The kernel window size with both character and word tokens is $h=3$ tokens. For word-level, the kernel of the last local max-pooling is 8 while it is equal 3 for char-level (because the size of the sequence is shorter). Following~\cite{Conneau-et-al:2016}, we experiment with two most effective configurations for word and character-level: $N_b=(4-4-4-4)$ and $N_b=(10-10-4-4)$, which are the number of convolutional layers in each of the four blocks.

\section{Experimental evaluation}
\label{sec:xp}

\begin{table*}[h]
\small
\centering
\begin{tabular}{|c|c|c|c|c|c|}
\hline \bf Models & \bf AGNews & \bf Yelp Bin & \bf Yelp Full & \bf DBPedia & \bf Yahoo  \\ \hline
Char shallow-and-wide CNN & 90.7 & 94.4 & 60.3 & 98.0 & 70.2  \\
Char-DenseNet $N_b=(4-4-4-4)$ Global Average-Pooling & 90.4 & 94.2 & 61.1 & 97.7 & 68.8 \\
Char-DenseNet $N_b=(10-10-4-4)$ Global Average-Pooling & 90.6 & 94.9 & 62.1 & 98.2 & 70.5 \\
Char-DenseNet $N_b=(4-4-4-4)$ Local Max-Pooling & 90.5 & 95.0 & 63.6 & 98.5 & 72.9 \\
Char-DenseNet $N_b=(10-10-4-4)$ Local Max-Pooling & 92.1 & 95.0 & 64.1 & 98.5 & 73.4 \\
Word shallow-and-wide CNN & 92.2 & \bf 95.9 & \bf 64.9 & \bf 98.7 & 73.0 \\
Word-DenseNet $N_b=(4-4-4-4)$ Global Average-Pooling & 91.7 & 95.8 & 64.5 & \bf 98.7 & 70.4* \\
Word-DenseNet $N_b=(10-10-4-4)$ Global Average-Pooling & 91.4 & 95.5 & 63.6 & 98.6 & 70.2* \\
Word-DenseNet $N_b=(4-4-4-4)$ Local Max-Pooling & 90.9 & 95.4 & 63.0 & 98.0 & 67.6* \\
Word-DenseNet $N_b=(10-10-4-4)$ Local Max-Pooling & 88.8 & 95.0 & 62.2 & 97.3 & 68.4* \\
\hline
bag of words~\cite{Zhang-et-al:2015} & 88.8 & 92.2 & 58.0 & 96.6 & 68.9 \\
ngrams~\cite{Zhang-et-al:2015} & 92.0 & 95.6 & 56.3 & 98.6 & 68.5 \\
ngrams TFIDF~\cite{Zhang-et-al:2015} & 92.4 & 95.4 & 54.8 & \bf 98.7 & 68.5 \\
fastText~\cite{Joulin-et-al:2016} & \bf 92.5 & 95.7 & 63.9 & 98.6 & 72.3 \\
char-CNN~\cite{Zhang-et-al:2015} & 87.2 & 94.7 & 62.0 & 98.3 & 71.2 \\
char-CRNN~\cite{Xiao-and-Cho:2016} & 91.4 & 94.5 & 61.8 & 98.6 & 71.7 \\
very deep char-CNN~\cite{Conneau-et-al:2016} & 91.3 & 95.7 & 64.7 & \bf 98.7 & 73.4 \\
Naive Bayes~\cite{Yogatama-et-al:2017} & 90.0 & 86.0 & 51.4 & 96.0 & 68.7 \\
Kneser-Ney Bayes~\cite{Yogatama-et-al:2017} & 89.3 & 81.8 & 41.7 & 95.4 & 69.3 \\
MLP Naive Bayes~\cite{Yogatama-et-al:2017} & 89.9 & 73.6 & 40.4 & 87.2 & 60.6 \\
Discriminative LSTM~\cite{Yogatama-et-al:2017} & 92.1 & 92.6 & 59.6 & \bf 98.7 & \bf 73.7 \\
Generative LSTM-independent comp.~\cite{Yogatama-et-al:2017} & 90.7 & 90.0 & 51.9 & 94.8 & 70.5 \\
Generative LSTM-shared comp.~\cite{Yogatama-et-al:2017} & 90.6 & 88.2 & 52.7 & 95.4 & 69.3 \\
\hline
\end{tabular}
\caption{Accuracy of our proposed models (10 top rows) and of state-of-the-art models from the litterature (13 bottom rows).}
\label{tab:res}
\end{table*}

\subsection{Tasks and data}

We test our models on the 5 datasets used in~\cite{Zhang-et-al:2015} and summarized in Table~\ref{tab:corpus}.
These datasets are:
\begin{itemize}
\item AGNews: internet news articles~\cite{AGNews} composed of titles plus descriptions and classified into 4 categories: World, Entertainment, Sports and Business, with 30k training samples and 1.9k test samples per class.
\item Yelp Review Polarity: The Yelp review dataset is obtained from the Yelp Dataset Challenge in 2015. Each polarity dataset has 280k training samples and 19k test samples.
\item Yelp Review Full: The Yelp review dataset is obtained from the Yelp Dataset Challenge in 2015. It has four polarity star labels: 1 and 2 as negative, and 3 and 4 as positive. Each star label has 130k training samples and 10k testing samples.
\item DBPedia: DBPedia is a 14 non-overlapping classes picked from DBpedia 2014 (wikipedia). Each class has 40k training samples and 5k testing samples.
\item Yahoo! Answers: ten largest main categories from Yahoo! Answers Comprehensive Questions and Answers version 1.0. Each class contains 140k training samples and 5k testing samples, including question title, question content and best answer. For DenseNet on word-level, we only used 560k samples because of lack of memory.
\end{itemize}

\begin{table}[h!]
\small
\centering
\begin{tabular}{|l|r|r|r|l|}
\hline \bf Dataset & \bf \#y & \bf \#train & \bf \#test & \bf Task \\ \hline
AGNews & 4 & 120k & 7.6k  & ENC \\
Yelp Binary & 2 & 560k & 38k & SA \\
Yelp Full & 5 & 650k & 38k & SA \\
DBPedia & 14 & 560k & 70k & OC \\
Yahoo & 10 & 1 400k & 60k & TC \\
\hline
\end{tabular}
\caption{Statistics of datasets used in our experiments: number of training tokens ({\bf \#train}), of test tokens ({\bf \#test}) and of target labels ({\bf \#y}); \textbf{ENC}: English News Categorization. \textbf{SA}: Sentiment Analysis, \textbf{OC}: Ontology Classification, \textbf{TC}: Topic Classification}
\label{tab:corpus}
\end{table}

\subsection{Hyperparameters and Training}

For all experiments, we train our model's parameters with the Adam Optimizer \cite{Kingma-and-Ba:2014} with an initial learning rate of 0.001, a mini-batch size of 128. The model is implemented using Tensorflow and is trained on a GPU cluster (with 12Gb RAM on GPU). The hyperparameters are chosen following~\cite{Zhang-et-al:2015} and~\cite{Kim:2014}, which are described below. On average, it takes about 10 epochs to converge.

\subsubsection{Character-level}

Following \cite{Zhang-et-al:2015}, each character is represented as a one-hot encoding vector where the dictionary contains the following 69 tokens: $"abcdefghijklmnopqrstuvwxyz0123456789-,$ $;.!?:’"/|\_\#\%\hat{}\&\star\tilde{}'+=<>()[]{}"$. The maximum sequence length is 1014 following~\cite{Zhang-et-al:2015}; smaller texts are padded with 0 while larger texts are truncated.
The convolutional layers are initialized following~\cite{Glorot-and-Bengio:2010}.

\subsubsection{Word-level}

The embedding matrix \textit{W} is initialized randomly with the uniform distribution between $[-0.1;0.1]$ and is updated during model's training using backpropagation. The embedding vectors have 300 dimensions and are initialized with word2vec vectors pretrained on 100 billion words from Google News~\cite{Mikolov-et-al:2013}. Out-of-vocabulary words are initialized randomly. A dropout of 0.5 is used on shallow model to prevent overfitting.

The shallow-and-wide CNN requires 10 hours of training on the smallest dataset, and one day on the largest.
The DenseNet respectively requires 2 and 4 days for training.

\subsection{Experimental results}

Table~\ref{tab:res} details the accuracy obtained with our models (10 rows on top) and compare them with state-of-the-art results (13 rows at the bottom)
on 5 corpus and text classification tasks (columns).
The models from the litterature we compare to are:
\begin{itemize}
\item \textbf{bag of words:} The BOW model is based on the most frequent words from the training data~\cite{Zhang-et-al:2015}
\item \textbf{ngrams:} The bag-of-ngrams model exploits the most frequent word n-grams from the training data~\cite{Zhang-et-al:2015}
\item \textbf{ngrams TFIDF:} Same as the ngrams model but uses the words TFIDF (term-frequency inverse-document-frequency) as features~\cite{Zhang-et-al:2015}
\item \textbf{fastText:} A linear word-level model with a rank constraint and fast loss approximation~\cite{Joulin-et-al:2016}
\item \textbf{char-CNN:} Character-level Convolutional Network with 6 \textit{hand-designed} CNN layers~\cite{Zhang-et-al:2015}
\item \textbf{char-CRNN:} Recurrent Layer added on top of a Character Convolutional Network~\cite{Xiao-and-Cho:2016}
\item \textbf{very deep CNN:} Character-level model with 29 Convolutional Layers inspired by ResNet~\cite{Conneau-et-al:2016}
\item \textbf{Naive Bayes:} A simple count-based word unigram language model based on the Naive Bayes assumption~\cite{Yogatama-et-al:2017}
\item \textbf{Kneser-Ney Bayes:} A more sophisticated word count-based language model that uses tri-grams and Kneser-Ney smoothing~\cite{Yogatama-et-al:2017}
\item \textbf{MLP Naive Bayes:} An extension of the Naive Bayes word-level baseline using a two layer feedforward neural network~\cite{Yogatama-et-al:2017}
\item \textbf{Discriminative LSTM:} Word-level model with logistic regression on top of a traditional LSTM~\cite{Yogatama-et-al:2017}
\item \textbf{Generative LSTM-independent comp.:} A class-based word language model with no shared parameters across classes~\cite{Yogatama-et-al:2017}
\item \textbf{Generative LSTM-shared comp.:} A class-based word language model with shared components across classes~\cite{Yogatama-et-al:2017}
\end{itemize}

\begin{figure*}[h]
\centering
\includegraphics[scale=0.7]{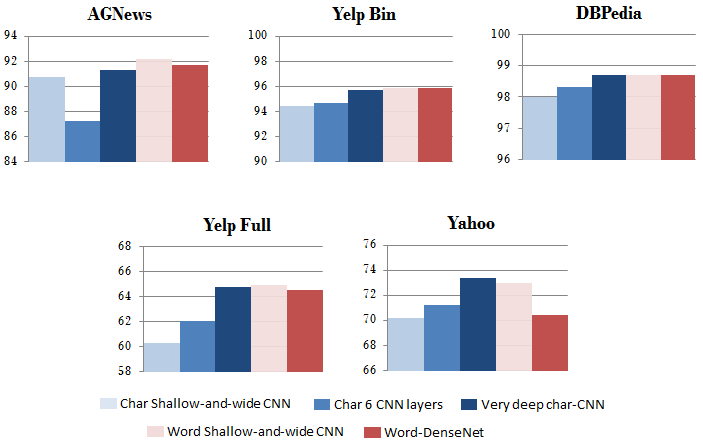}
\caption{Comparison of character (\textit{in blue, on the left}) and word-level (\textit{in red, on the right}) models on all datasets. On character-level, we compare our shallow-and-wide model with the 6 CNN layers of~\cite{Zhang-et-al:2015} and the 29-layers CNN of~\cite{Conneau-et-al:2016}. On word-level, we compare the shallow-and-wide CNN with our proposed DenseNet.}
\label{fig:res}
\end{figure*}

Figure~\ref{fig:res} visually compares the performances of 3 character-level models with 2 word-level models. Character-level models include: our shallow-and-wide CNN model with two models on the litterature 6 CNN layers~\cite{Zhang-et-al:2015}, 29 CNN layers~\cite{Conneau-et-al:2016}. On word-level, we present our shallow-and-wide CNN with the best DenseNet $N_b=(4-4-4-4)$ using Global Average-Pooling.

The main conclusions of these experiments are threefold:

\subsubsection*{Impact of depth for character-level models}
Deep character-level models do not significantly outperform the shallow-and-wide network. A shallow-and-wide network (row 1 in Table~\ref{tab:res}) achieves 90.7\%, 94.4\%, 98.0\% on AGNews, Yelp Bin, DBPedia respectively, comparing to  91.3\%, 95.7\%, 98.7\% of a very deep CNN~\cite{Conneau-et-al:2016}. Although the deep structure achieves a slight gain in performance on these three datasets, the difference is not significant.
Interestingly, a very simple shallow-and-wide CNN can get very close results to the deep 6 CNN layers of~\cite{Zhang-et-al:2015} which structure must be designed meticulously.

For the smallest dataset AGNews, we suspect that the deep model {\bf char-CNN} performs badly because it needs more data to take benefit from depth.
The deep structure gives an improvement of about 4\% on Yelp Full and Yahoo (first row of Table~\ref{tab:res} vs. {\bf very deep char CNN}), which is interesting but does not match the gains observed in image classification.
We have tried various configurations: $N_f=(15,20,25)$, $N_f=(10,15,20,25)$ and $N_f=(15,22,29,36)$ on shallow models but they didn't do better.

\subsubsection*{Impact of depth for word-level models}

The DenseNet is better with 20 layers $N_b=(4-4-4-4)$ than with 32 layers $N_b=(10-10-4-4)$ and Global Average-Pooling is better than the traditional Local Max-pooling. It is the opposite to char-level. This is likely a consequence of the fact that the observed sequence length is much shorted with words than with characters.

However, the main striking observation is that all deep models are matched or outperformed by the shallow-and-wide model on all datasets,
although it is still unclear whether this is because the input sequences are too short to benefit from depth or for another reason.
Further experiments are required to investigate the underlying reasons of this failure of depth at word-level.

\subsubsection*{State-of-the-art performances with shallow-and-wide word-level model}

With a shallow-and-wide network on word-level, we achieved a very close state-of-the-art (SOTA) result on AGNews, SOTA on 2 datasets DBPedia, Yahoo and set new SOTA on 2 datasets: Yelp Binary and Yelp Full.
We also empirically found that a word-level shallow model may outperform a very deep char-level network.
This confirms that word observations are still more effective than character inputs for text classification.
In practice, quick training on word-level with a simple convolutional model may already produce good result.

\subsection{Discussion}

\subsubsection*{Text representation - discrete, sparse}
\begin{figure*}[h!]
\centering
\includegraphics[scale=0.7]{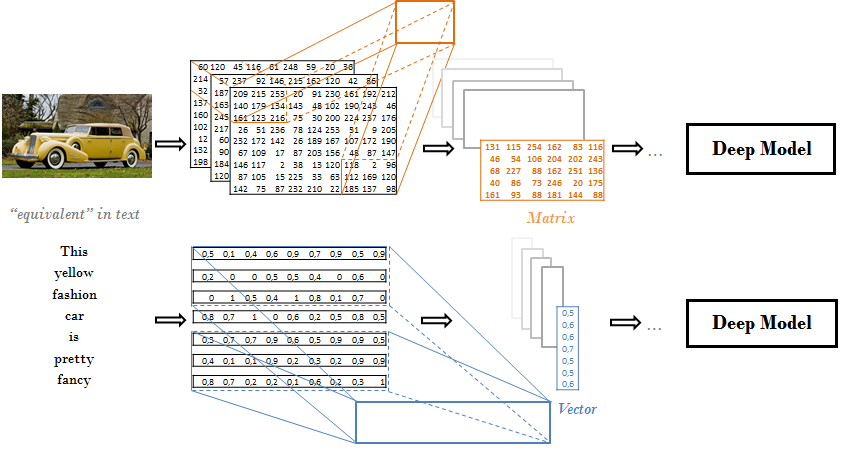}
\caption{Inside operations of convolution on \textbf{Image} vs \textbf{Text}. \textbf{Image} has real-valued and dense. \textbf{Text} has discrete tokens, many artificial and sparse values representation. The output of the first convolution layer on image is still a \textit{Matrix} but on text, it is reduced to a \textit{Vector}. }
\label{Image_vs_text}
\end{figure*}

Very deep models do not seem to bring a significant advantage over shallow networks for text classification, as opposed to
their performances in other domains such as image processing.
We believe one possible reason may be related to the fact that images are represented as real and dense values, as opposed to discrete, artificial and sparse representation of text. The first convolutional operation results in a matrix (2D) for image while it results in a vector (1D) for text (see Figure \ref{Image_vs_text}). The same deep network applied to two different representations (dense and sparse) will obviously get different results. Empirically, we found that a deep network on 1D (text) is less effective and learns less information than on 2D (image).

\noindent{\bf Local vs Global max-pooling}

A global max-pooling \cite{Collobert-and-Weston:2008}, which retrieves the most influencial feature could already be good enough for sparse and discrete input text,
and gives similar results than a local max-pooling with a deep network.

\noindent{\bf Word vs Character level}

Char-level could be a choice but word-level is still the most effective method. Moreover, in order to use char-level representation, we \textit{must} use a very deep model, which is less practical because it 
takes a long time to train.

\section{Conclusion}
In computer vision, several works have shown the importance of depth in neural networks and the major benefits that can be gained by
stacking many well-designed convolutional layers in terms of performances.
However, such advantages do not necessarily transfer to other domains, and in particular Natural Language Processing,
where the impact of depth in the model is still unclear.
This work exploits a number of additional experiments to further explore this question and potentially bring some new insights or
confirm previous findings.
We further investigate another related question about which type of textual inputs, characters or words, should be chosen at a given depth.
By evaluating on several text classification and sentiment analysis tasks, we show that a shallow-and-wide convolutional neural
network at the word-level is still the most effective, and that increasing the depth of such convolutional models with word inputs
does not bring significant improvement.
Conversely, deep models outperform shallow networks when the input text is encoded as a sequence of characters, but although such deep models
approach the performances of word-level networks, they are still worse on the average.
Another contribution of this work is the proposal of a new deep model that is an adaptation of DenseNet for text inputs.

Based on the litterature and the results presented in this work, our main conclusion is that deep models have not yet proven to
be more effective than shallow models for text classification tasks.
Nevertheless, further researches should be realized to confirm or infirm this observation on other datasets, 
natural language processing tasks and models.
Indeed, this work derives from reference deep models that have originally been developed for image processing,
but novel deep architectures for text processing might of course challenge this conclusion in the near future.

\section*{Acknowledgments}
Part of the experiments realized in this work have been done on two GPU clusters: \textit{Grid5000 Inria/Loria Nancy, France} and \textit{Romeo Reims, France}. We would like to thank both consortiums for giving us access to their resources.

\bibliographystyle{emnlp_natbib}
\bibliography{emnlp2017}

\end{document}